\documentclass[conference]{IEEEtran}

\usepackage{etoolbox}
\makeatletter
\patchcmd{\@makecaption}
  {\scshape}
  {}
  {}
  {}
\makeatletter
\patchcmd{\@makecaption}
  {\\}
  {.\ }
  {}
  {}
\makeatother
\def\tablename{Table}

\IEEEoverridecommandlockouts
\usepackage{cite}
\usepackage{amsmath,amssymb,amsfonts}
\usepackage{algorithmic}
\usepackage{graphicx}
\usepackage{textcomp}
\usepackage{xcolor}
\usepackage{subfigure}
\usepackage{multirow}
\usepackage{bbding}
\usepackage{enumitem}
\usepackage[T1]{fontenc}

\def\BibTeX{{\rm B\kern-.05em{\sc i\kern-.025em b}\kern-.08em
    T\kern-.1667em\lower.7ex\hbox{E}\kern-.125emX}}
    
\begin{document}

\newcommand \eg {e.g.\ }
\newcommand \ie {i.e.\ }
\newcommand \beq {\begin{equation}}
\newcommand \eeq {\end{equation}}

\title{DEAP-3DSAM: Decoder Enhanced and Auto Prompt SAM for 3D Medical Image Segmentation\\
}

\author{\IEEEauthorblockN{Fangda Chen$^*$ \thanks{ * Equal contribution.}}
\IEEEauthorblockA{\textit{College of Computer Science and Tech.} \\
\textit{National University of Defense Tech.}\\
Changsha, China \\
fdchen.nudt@nudt.edu.cn}
\and
\IEEEauthorblockN{Jintao Tang$^{*\dagger}$ \thanks{ $\dagger$ Corresponding~author: Jintao~Tang, Ting~Deng.}}
\IEEEauthorblockA{\textit{College of Computer Science and Tech.} \\
\textit{National University of Defense Tech.}\\
Changsha, China \\
tangjintao@nudt.edu.cn}
\and
\IEEEauthorblockN{Pancheng Wang}
\IEEEauthorblockA{\textit{College of Computer Science and Tech.} \\
\textit{National University of Defense Tech.}\\
Changsha, China \\
wangpancheng13@nudt.edu.cn}
\and
\IEEEauthorblockN{Ting Wang}
\IEEEauthorblockA{\textit{College of Computer Science and Tech.} \\
\textit{National University of Defense Tech.}\\
Changsha, China \\
tingwang@nudt.edu.cn}
\and
\IEEEauthorblockN{Shasha Li}
\IEEEauthorblockA{\textit{College of Computer Science and Tech.} \\
\textit{National University of Defense Tech.}\\
Changsha, China \\
shashali@nudt.edu.cn}
\and
\IEEEauthorblockN{Ting Deng$^\dagger$}
\IEEEauthorblockA{\textit{Department of GI Medical Onco.} \\
\textit{TMUCIH, NCRCC}\\
Tianjin, China \\
xymcdengting@126.com}
}

\maketitle

\begin{abstract}
The Segment Anything Model (SAM) has recently demonstrated significant potential in medical image segmentation. Although SAM is primarily trained on 2D images, attempts have been made to apply it to 3D medical image segmentation. However, the pseudo 3D processing used to adapt SAM results in spatial feature loss, limiting its performance. Additionally, most SAM-based methods still rely on manual prompts, which are challenging to implement in real-world scenarios and require extensive external expert knowledge. To address these limitations, we introduce the \textbf{D}ecoder \textbf{E}nhanced and \textbf{A}uto \textbf{P}rompt \textbf{SAM} (\textbf{DEAP-3DSAM}) to tackle these limitations. Specifically, we propose a Feature Enhanced Decoder that fuses the original image features with rich and detailed spatial information to enhance spatial features. We also design a Dual Attention Prompter to automatically obtain prompt information through Spatial Attention and Channel Attention. We conduct comprehensive experiments on four public abdominal tumor segmentation datasets. The results indicate that our DEAP-3DSAM achieves state-of-the-art performance in 3D image segmentation, outperforming or matching existing manual prompt methods. Furthermore, both quantitative and qualitative ablation studies confirm the effectiveness of our proposed modules.
\end{abstract}


\section{Introduction}
3D medical image segmentation delineates lesions to assist doctors in disease diagnosis and plays a crucial role in modern medicine. As deep learning technology advances, 3D medical image segmentation techniques have achieved significant success. Early methods utilized the structures of Convolutional Neural Network (CNN), such as UNet \cite{unet} and nnU-Net \cite{nnunet}. CNN-based approaches extract local features from 3D medical images and capture high-dimensional features by stacking layers. However, they primarily focus on local features and cannot often capture global features. In contrast, Transformer architectures, based on self-attention, excel at capturing global features. Consequently, researchers have proposed several Transformer-based models, including UNETR \cite{unetr}, Swin UNETR \cite{swinunetr}, and UNETR++ \cite{unetr++}. These models have demonstrated remarkable success in routine tasks, such as organ segmentation and brain tumor segmentation. Nevertheless, they may encounter challenges when applied to more complex segmentation tasks.

\begin{figure}[t]
  \centering
  \includegraphics[width=0.5\textwidth]{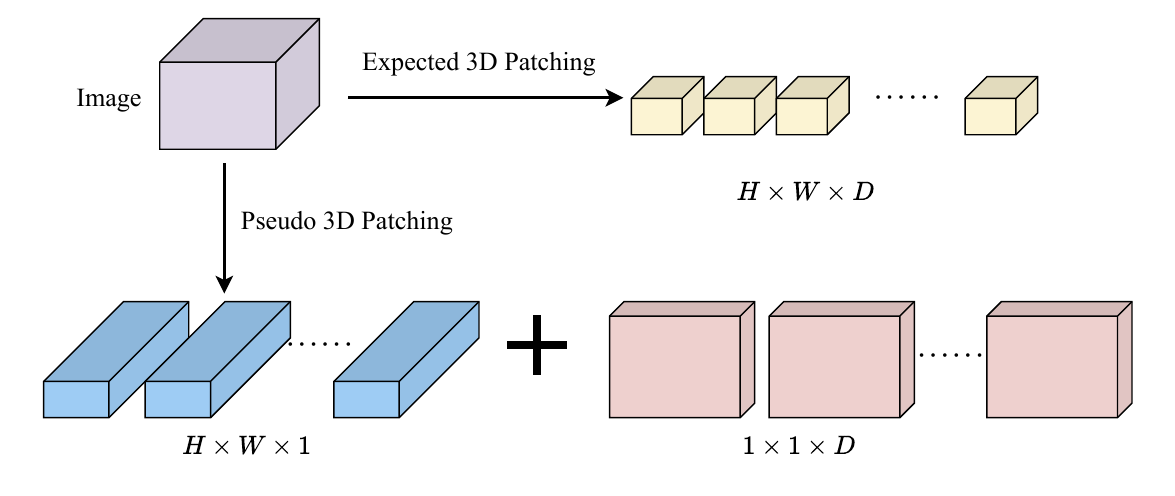}
  \caption{An illustration of pseudo 3D patching and expected 3D patching.}
  \label{pesudo 3d patching}
\end{figure}

Recently, Meta introduces a pre-trained model for 2D image segmentation known as the Segment Anything Model (SAM) \cite{sam}. SAM has been trained on over 11 million general images and 1.1 billion masks, supporting four types of human prompts, \ie points, boxes, masks and text. SAM demonstrates exceptional performance across various segmentation tasks, maintaining efficacy in both zero-shot and few-shot scenarios. Motivated by its powerful capabilities, researchers have begun exploring its applications in 3D medical images. A straightforward approach involves slicing a single 3D image into multiple 2D slices, \eg SAMed \cite{samed} and nnSAM \cite{nnsam}. However, this method results in a significant increase in processing time and a loss of depth information. Alternative approaches involve integrating entire 3D images into SAM using pseudo-3D processing techniques, \eg Medical-SAM-Adapter \cite{msa} and 3DSAM-Adapter \cite{3dsamadapter}.

Leveraging the robust capabilities of the Segment Anything Model (SAM), SAM-based approaches have demonstrated performance that is comparable to, or even surpasses, traditional 3D medical image segmentation techniques \cite{samed, nnsam, msa, 3dsamadapter}. Despite these advancements, SAM-based methods encounter specific challenges, including: (i) \textbf{Spatial feature loss in pseudo 3D processing.} A typical pseudo 3D processing method is shown in Fig.~\ref{pesudo 3d patching}, \ie pseudo 3D patching \cite{3dsamadapter}. As shown, pseudo 3D processing approximates the expected 3D image processing. However, this approximation results in an incomplete representation of spatial information, leading to the loss of spatial features. This deficiency manifests as issues such as inaccurate segmentation boundaries and inadequately delineated target regions, which is particularly detrimental in complex medical scenarios. For example, in the case of abdominal tumors, precise delineation of tumor boundaries and sizes is critical for accurate disease assessment. Consequently, the loss of spatial features limits the effectiveness of existing methods in complex 3D medical image segmentation scenarios. (ii) \textbf{The limited practicality of SAM-based methods with manual prompts.} SAM is proposed for 2D image segmentation and supports various prompts, including foreground points, background points, bounding boxes, and masks. However, applying these interactive segmentation techniques often presents significant challenges in real-world 3D image segmentation contexts. On the one hand, manual prompting necessitates a sophisticated 3D annotation tool designed to facilitate expert interaction. Implementing such tools becomes particularly challenging when dealing with complex 3D prompts, such as boxes and masks. On the other hand, manual prompting requires precise expert knowledge, necessitating manual annotation by multiple specialists. This process is not only time-consuming but also labor-intensive. Therefore, there is an urgent need to develop an automated prompting mechanism for SAM in the context of 3D medical image segmentation.

To address the limitations above, we propose a novel SAM-based framework for 3D medical image segmentation, \textbf{D}ecoder \textbf{E}nhanced and \textbf{A}uto \textbf{P}rompt \textbf{SAM}, called \textbf{DEAP-3DSAM}. The efficacy of DEAP-3DSAM is primarily derived from two innovative modules. (i) \textbf{Feature Enhanced Decoder}: To mitigate the issue of spatial feature loss, we have designed a Feature Enhanced Decoder that strengthens the SAM-encoded feature maps by integrating the original image features. We consider that the original image contains complete spatial information, including precise sizes and intricate texture details. Therefore, we extract the spatial features of the original image using convolutional networks and incorporate them into the SAM-encoded feature maps. Additionally, drawing inspiration from the success of UNet \cite{unet} in medical image segmentation, we integrate the original image features into the four feature maps at varying depths. (ii) \textbf{Dual Attention Prompter}: Inspired by 2D image segmentation prompt techniques, we recognize that these manual prompts provide prior location information essential for localizing the segmentation region. This prior information enables the model to focus on the region of interest. We argue that these approaches can be replaced with a spatial self-attention mechanism. However, the computational cost of global self-attention on 3D medical images is prohibitive. To address this, we incorporate the linear self-attention mechanism from Linformer \cite{linformer} as an efficient approximation, denoted as Spatial Attention. Furthermore, influenced by the Dual Attention approach \cite{dan}, we propose that channel information can also serve as automatic prompt information. The channel self-attention assists the model in identifying the most significant channels in the feature maps, thereby enhancing performance. Consequently, we introduce Channel Attention in conjunction with Spatial Attention to establish dual attention.

In conclusion, the main contributions of this paper are as follows:
\begin{itemize}[leftmargin=*]
\item We have innovatively designed a 3D mask decoder for SAM, namely Feature Enhanced Decoder, which incorporates the original image information to mitigate the spatial feature loss due to pseudo 3D processing.
\item We propose an automatic prompting method called the Dual Attention Prompter, which utilizes both Spatial Attention and Channel Attention to enhance the suitability of SAM for practical applications in 3D medical image segmentation.
\item Through extensive experimentation on four datasets of abdominal tumors, our DEAP-3DSAM, which operates without manual prompting, has demonstrated superior performance, matching or surpassing current manual prompt-driven methodologies. Notably, it outperformed the 3DSAM-adapter by 4.62\% for KiTS21, 0.87\% for Pancreas, 10.34\% for LiTS17, and 8.09\% for Colon.
\end{itemize}

\begin{figure*}[htbp]
  \centering
  \includegraphics[width=.8\textwidth]{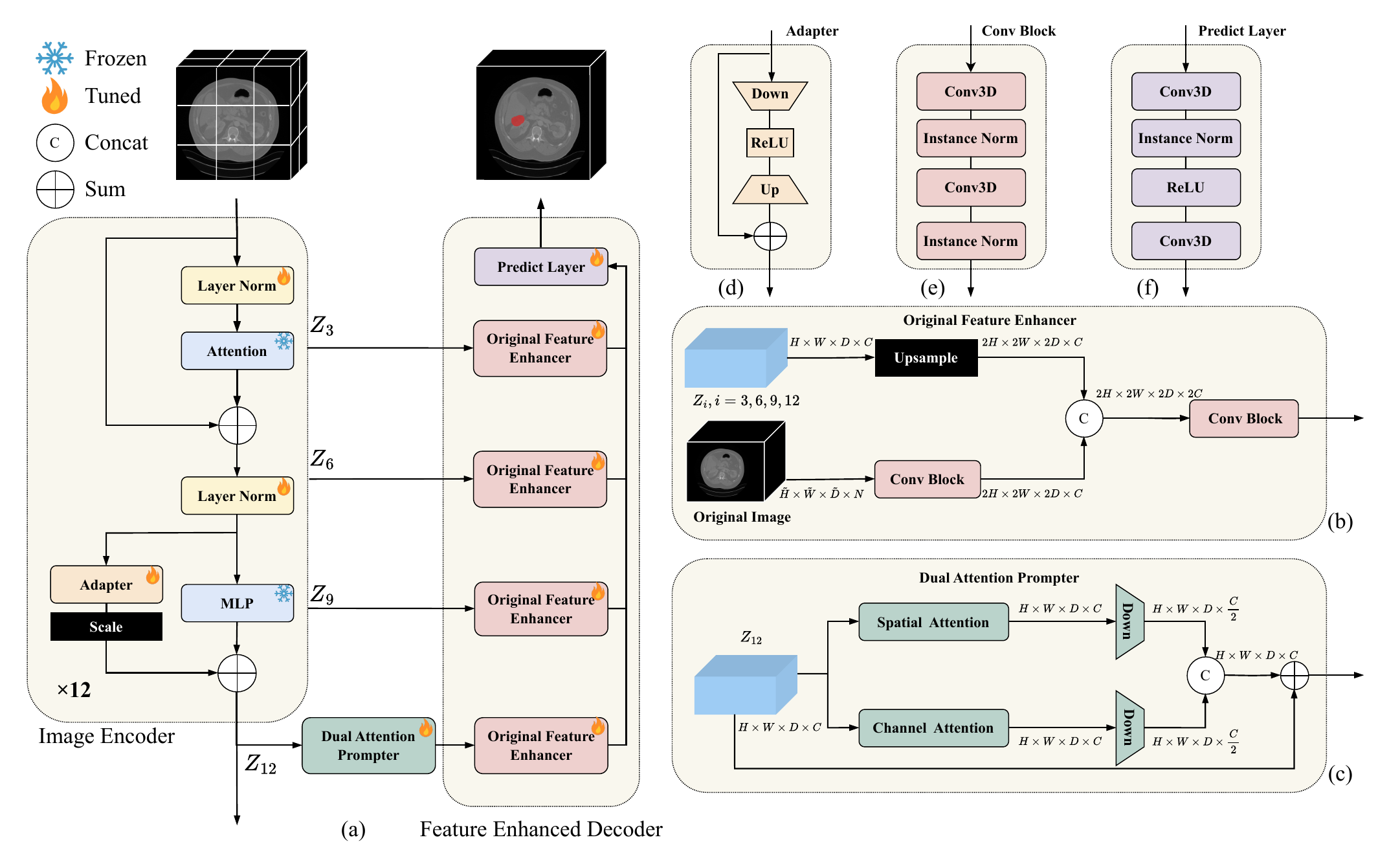} 
  \caption{The overall framework of our DEAP-3DSAM. (a) represents the overall processes involved. The 3D medical image is initially decomposed into a series of patches through a pseudo 3D patching process. These patches are subsequently processed by the Image Encoder. The feature map from the final transformer layer, denoted as $z_{12}$, must be processed by the Dual Attention Prompter. Subsequently, $z_{12}$, along with other intermediate feature maps, $z_3$, $z_6$, and $z_9$, is fed into the Feature Enhanced Decoder. The final segmentation predictions are generated by the Predict Layer within the Feature Enhanced Decoder. (b) illustrates the Feature Enhanced Decoder, primarily composed of Original Feature Enhancers. The Original Feature Enhancer merges the upsampled feature map with the original features from the input image, subsequently outputting the augmented features through a convolutional block. (c) demonstrates that the Dual Attention Prompter employs both Spatial and Channel Attention, followed by the concatenation of the resulting features for output. Finally, (d), (e), and (f) depict the structural components of the adapter, convolution block, and prediction layer, respectively.} 
  \label{framework} 
\end{figure*}

\section{Related Work}
In this section, we first review the conventional methodologies utilizing Transformers or CNN. Subsequently, we summarize the current state-of-the-art SAM-based models.

\subsection{3D Medical Image Segmentation with Transformer or CNN}
Transformer has become a pivotal technology in natural language processing and computer vision. Researchers are increasingly exploring the application of Transformer in medical image segmentation \cite{unetr, swinunetr, unetr++, nnformer}. UNETR \cite{unetr} proposes an architecture that unifies a Transformer-based encoder with a CNN-based decoder to effectively capture both global and local features. Swin UNETR \cite{swinunetr} incorporates the Swin Transformer in the encoder to enhance the accuracy of brain tumor segmentation. Additionally, other researchers have employed CNN-based methodologies \cite{nnunet, 3duxnet}. nnU-Net \cite{nnunet} utilizes adaptive data augmentation and a U-shaped CNN to achieve superior performance in various medical image segmentation tasks. 3D UX-Net \cite{3duxnet} adapts the Swin Transformer \cite{swin_transformer} with 3D deep convolutional layers for more efficient and effective segmentation. These 3D medical image segmentation methods have demonstrated success in applications such as organ segmentation and brain tumor segmentation. Furthermore, researchers have integrated Transformer and CNN to capitalize on the strengths of both approaches \cite{trans_uneter, sctransnet, transfuse}. Trans-UNeter \cite{trans_uneter} employs a CNN-based encoder combined with an attention-based decoder to explore the complex feature relationships between the encoder and decoder. SCTransNet \cite{sctransnet} utilizes a CNN for feature extraction and incorporates spatial and channel attention mechanisms within the Transformer framework. TransFuse \cite{transfuse} adopts an intuitive strategy by fusing the branches of the Transformer and CNN. However, segmentation remains less effective for some sophisticated scenarios, such as abdominal tumors, due to the intricate images and uncertain tumor shapes.

\subsection{3D Medical Image Segmentation based on SAM}
SAM is a pre-trained model renowned for its robust zero-shot and few-shot image segmentation capabilities. Many researchers are investigating the transfer of these exceptional segmentation abilities to 3D medical image segmentation \cite{samed, nnsam, msa, 3dsamadapter, 3dusam, sam_review}. However, SAM is initially designed for 2D images and requires adaptation to process 3D images effectively. Currently, there are two primary approaches to this adaptation. The first approach involves slicing a single 3D image into multiple 2D images and subsequently combining the predicted slices during post-processing \cite{samed, nnsam}. SAMed \cite{samed} employs LoRA \cite{lora} to fine-tune SAM, while nnSAM \cite{nnsam} enhances the image features of nnU-Net \cite{nnunet} using the SAM image encoder. The second method entails inputting a 3D image into the SAM model as a whole \cite{msa, 3dsamadapter, 3dusam}. The crux of these methods lies in modifying the SAM image encoder structure and reusing the pre-training parameters for 3D images. Medical-SAM-Adapter \cite{msa} models z-axis features by horizontally reusing the attention modules within the SAM image encoder. 3DSAM-Adapter \cite{3dsamadapter} patches from the sliding window along the z-axis with the original 2D sliding window, approximating pseudo 3D patching across the entire 3D image. 3D-U-SAM \cite{3dusam} introduces a specialized adapter to map 3D images to 2D images. The first approach results in a complete loss of depth semantics, and the training and inference times increase significantly due to the numerous slices of the 3D image. The second approach has achieved some success in specific 3D medical image segmentation tasks. However, these methods rely on coarse pseudo 3D processing to adapt SAM to 3D images, leading to greater spatial feature loss and suboptimal performance on challenging segmentation tasks. Furthermore, these methods still depend on manual prompts, which can be difficult to implement, and the effectiveness of these prompts can not be guaranteed.

\section{Method}
In this section, we describe the overall framework of the proposed DEAP-3DSAM. As illustrated in Fig. \ref{framework}, our DEAP-3DSAM framework comprises three primary components: an image encoder, a Feature Enhanced Decoder, and a Dual Attention Prompter.

\subsection{Image Encoder}
As illustrated in the left of Fig.~\ref{framework} (a), our DEAP-3DSAM utilizes pseudo 3D patching, similar to 3DSAM-Adapter \cite{3dsamadapter}. For a 3D image with shape of $\bar{H} \times \bar{W} \times \bar{D} \times N$, the dimension transforms to $H \times W \times D \times C$ after pseudo 3D patching and embedding. Here, $\bar{H}, \bar{W}, \bar{D}$ represent the initial height, width and depth of the image; $N$ denotes the number of channels; $H, W, D$ indicate the height, width and depth of the feature map; and $C$ is the channel dimension. The  feature map is then input into the image encoder. The image encoder comprises 12 Transformer layers, each containing an attention layer, a Multi-Layer Perceptron (MLP) layer and two normalization layers. To maximize the utilization of the pre-trained parameters of SAM, we fix the parameters of the attention and MLP layers and incorporate a Scale Parallel Adapter for fine-tuning. He et al. \cite{scalepa} propose the Scale Parallel Adapter for fine-tuning Transformer layers. They demonstrate that a parallel-placed adapter outperforms a sequentially-placed one and that a parallel-placed adapter to the Feed-Forward Network is more effective than one aligned with multi-head attention \cite{scalepa}. Consequently, our image encoder employs the Scale Parallel Adapter on the MLP layer, as depicted in Fig.~\ref{framework} (d), and the equation is calculated as follows,
\beq
Adapter(X) = ReLU(XW_{down})W_{up},
\eeq
where $X \in \mathbb{R}^{H \times W \times D \times C}$ represents the feature map; $ReLU(\cdot)$ denotes the activation function; both $W_{down} \in \mathbb{R}^{C \times l}$ and $W_{up} \in \mathbb{R}^{l \times C}$ denoterepresent the mapping weights, with $l$ is a smaller dimension.

In conclusion, the entire Transformer layer is computed as follows, and the output feature map of $i$-th Transformer layer is denoted as $Z_i \in \mathbb{R}^{H \times W \times D \times C}$,
\beq
\dot{Z}_i = Norm(Z_{i-1}), \\
\eeq
\beq
\hat{Z}_i = Z_{i-1} + Attention^*(\dot{Z}_i),
\eeq
\beq
\ddot{Z}_i = Norm(\hat{Z}_i),
\eeq
\beq
Z_i = MLP^*(\ddot{Z}_i) + s \cdot Adapter(\ddot{Z}_i),
\eeq
where $Norm(\cdot)$ denotes the normalization layer; $Attention^*(\cdot)$ and $MLP^*(\cdot)$ denotes the attention and MLP layers with frozen parameters; $s$ denotes the scale coefficient, and $\dot{Z}_i, \hat{Z}_i, \ddot{Z}_i$ are intermediate variables. 

To leverage image semantics at various depths, we input the feature maps from every third layer of the image encoder, specifically, $Z_3,Z_6,Z_9$ and $Z_{12}$, into the decoder.

\subsection{Feature Enhanced Decoder}
We design a customized 3D mask decoder for SAM, named Feature Enhanced Decoder, as illustrated on the right of Fig. \ref{framework} (a). The Feature Enhanced Decoder comprises four Original Feature Enhancers and a prediction layer. Each Original Feature Enhancer is divided into two distinct branches. The first branch is responsible for extracting original image features using a Convolutional Block (Conv Block), which includes 3D convolutional layers and instance normalization, as shown in Fig. \ref{framework} (e). The second branch processes the feature map, requiring upsampling to match the dimensions of the original image features. To integrate semantics from various layers, the enhanced features from all enhancers are concatenated in the prediction layer. Therefore, the upsampling ratio must be moderate, as a large scale would significantly increase the number of parameters. Here $2H, 2W, 2D$ is employed. Finally, the features from both branches are concatenated and integrated through a convolutional block. In summary, the Feature Enhanced Decoder is formulated as follows,
\beq
\hat{E}_j = Upsample(Z_i),
\eeq
\beq
\bar{E}_j = ConvBlock(I),
\eeq
\beq
E_j = ConvBlock(Concat(\hat{E}_j, \bar{E}_j)),
\eeq
\beq
Y = Predict(Concat(E_1,...,E_j,...,E_4)),
\eeq
where $Z_i$ represents the  feature map from the $i$-th layer of the image encoder, $i=3,6,9,12$; $I$ represents the original image; $E_j$ denotes the output of the $j$-th enhancer, $j=1,2,3,4$; $Y$ denotes the predicted segmentation; $Upsample(\cdot)$ denotes upsample; $ConvBlock(\cdot)$ denotes the convolutional block; $Concat(\cdot)$ denotes concatenating the features; $Predict(\cdot)$ denotes the predict layer; $\hat{E}_ j, \bar{E}_j$ denote intermediate variables.

\begin{table*}[!h]
    \centering
    \caption{Overall performance comparison on four tumor segmentation datasets. Numbers in \textbf{bold} signify the best performance, while \underline{underlined} numbers denote the second-best. DEAP-3DSAM\ddag ~utilizes partial weight sharing between Spatial Attention and Channel Attention, whereas DEAP-3DSAM\dag ~does not.}
    \resizebox{\textwidth}{!}{
    \begin{tabular}{c|cc|cc|cc|cc|c}
    \hline
        \multirow{2}*{\textbf{Models}} & \multicolumn{2}{c|}{\textbf{KiTS21}} & \multicolumn{2}{c|}{\textbf{Pancreas}} &  \multicolumn{2}{c|}{\textbf{LiTS17}} & \multicolumn{2}{c|}{\textbf{Colon}} &  \multirow{2}*{\textbf{\#Tuned Params}} \\  \cline{2-9}
        ~ & DICE $\uparrow$ & NSD $\uparrow$ & DICE $\uparrow$ & NSD $\uparrow$ & DICE $\uparrow$ & NSD $\uparrow$ & DICE $\uparrow$ & NSD $\uparrow$ & ~ \\ \hline
        nnU-Net (Nat. Methods 2021) \cite{nnunet} & 73.07 & 77.47 & 41.65 & 62.54 & \textbf{60.10} & \textbf{75.41} & 43.91 & 52.52 & 30.76M \\ 
        TransBTS (MICCAI 2021) \cite{transbts} & 40.79 & 37.74 & 31.90 & 41.62 & 34.69 & 49.47 & 17.05 & 21.63 & 32.33M \\ 
        nnFormer (arXiv 2021) \cite{nnformer} & 45.14 & 42.28 & 36.53 & 53.97 & 45.54 & 60.67 & 24.28 & 32.19 & 149.49M \\ 
        Swin-UNETR (CVPR 2022) \cite{swinunetr} & 65.54 & 72.04 & 40.57 & 60.05 & 50.26 & 64.32 & 35.21 & 42.94 & 62.19M \\ 
        UNETR++ (arXiv 2022) \cite{unetr++} & 56.49 & 60.04 & 37.25 & 53.59 & 37.13 & 51.99 & 25.36 & 30.68 & 55.70M \\ 
        3D UX-Net (ICLR 2023) \cite{3duxnet} & 57.59 & 58.55 & 34.83 & 52.56 & 45.54 & 60.67 & 28.50 & 32.73 & 53.01M \\ \hline
        SAM-B (10 pts/slice) \cite{sam} & 40.07 & 34.96 & 30.55 & 32.91 & 8.56 & 5.97 & 39.14 & 42.70 & - \\ \hline
        3DSAM-Adapter (1 pt/volume) \cite{3dsamadapter} & 77.54 & 84.69 & 58.12 & 78.40 & 49.08 & 61.16 & 60.54 & 76.09 & 25.46M \\ 
        3DSAM-Adapter (3 pts/volume) \cite{3dsamadapter} & 80.44 & 87.68 & \underline{59.25} & \underline{79.56} & 53.74 & \underline{66.46} & 62.67 & 77.85 & 25.46M \\ 
        3DSAM-Adapter (10 pts/volume) \cite{3dsamadapter} & 80.59 & 88.18 & \textbf{59.81} & \textbf{80.16} & 52.97 & 66.18 & 63.36 & 79.02 & 25.46M \\ \hline
        DEAP-3DSAM\dag ~(Ours)  & \textbf{81.53} & \textbf{88.93} & 57.89 & 79.41 & 49.89 & 61.55 & \underline{65.16} & \underline{79.95} & 28.62M \\ 
        DEAP-3DSAM\ddag ~(Ours)  & \underline{81.12} & \underline{88.20} & 58.62 & 78.91 & \underline{54.16} & 65.71 & \textbf{65.43} & \textbf{80.21} & 28.48M \\ \hline
    \end{tabular}
    \label{overall performance}
    }
\end{table*}

\begin{figure*}[t]
	\centering
        \begin{minipage}{\textwidth}
            \centering
            \rotatebox{90}{\scriptsize{~KiTS21-Case 58}}
            \subfigure{
                \begin{minipage}[t]{0.12\linewidth}
                    \centering
                    \includegraphics[width=1\linewidth]{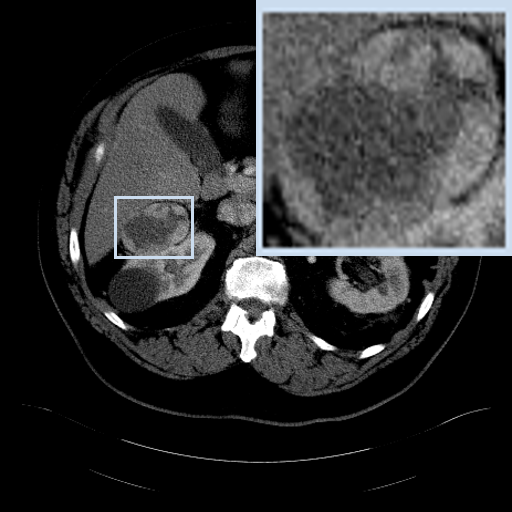}
                \end{minipage}
            }
            \subfigure{
                \begin{minipage}[t]{0.12\linewidth}
                    \centering
                    \includegraphics[width=1\linewidth]{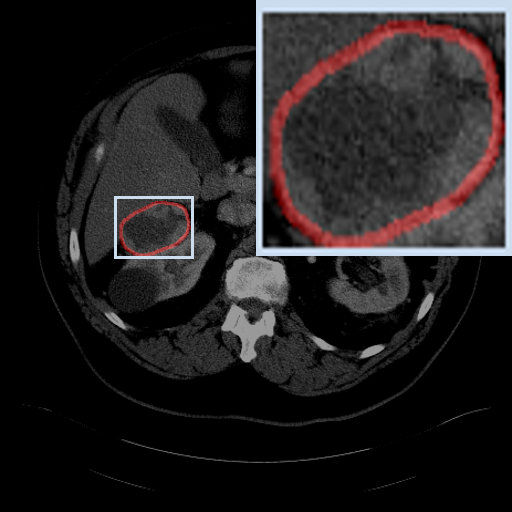}
                \end{minipage}
            }
            \subfigure{
                \begin{minipage}[t]{0.12\linewidth}
                    \centering
                    \includegraphics[width=1\linewidth]{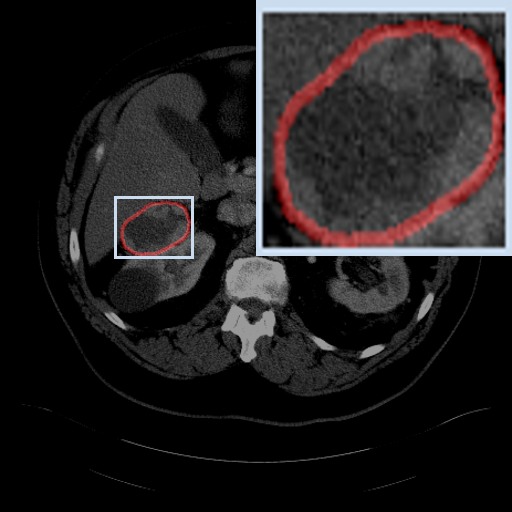}
                \end{minipage}
            }
            \subfigure{
                \begin{minipage}[t]{0.12\linewidth}
                    \centering
                    \includegraphics[width=1\linewidth]{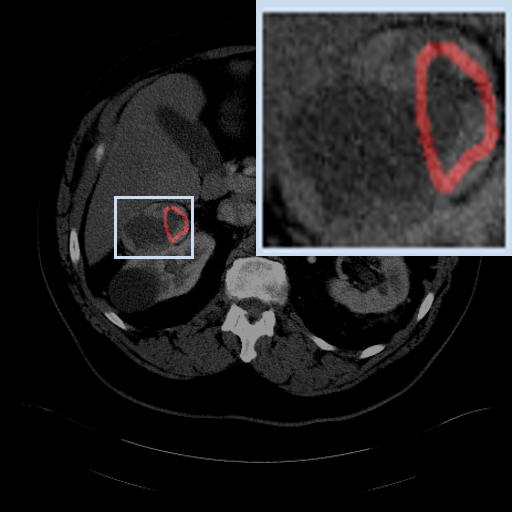}
                \end{minipage}
            }
            \subfigure{
                \begin{minipage}[t]{0.12\linewidth}
                    \centering
                    \includegraphics[width=1\linewidth]{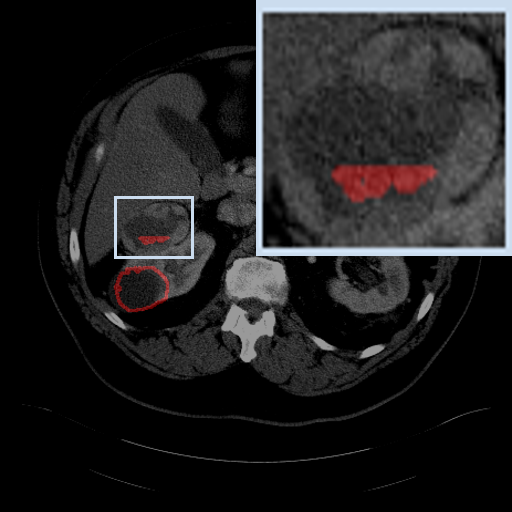}
                \end{minipage}
            }
            \subfigure{
                \begin{minipage}[t]{0.12\linewidth}
                    \centering
                    \includegraphics[width=1\linewidth]{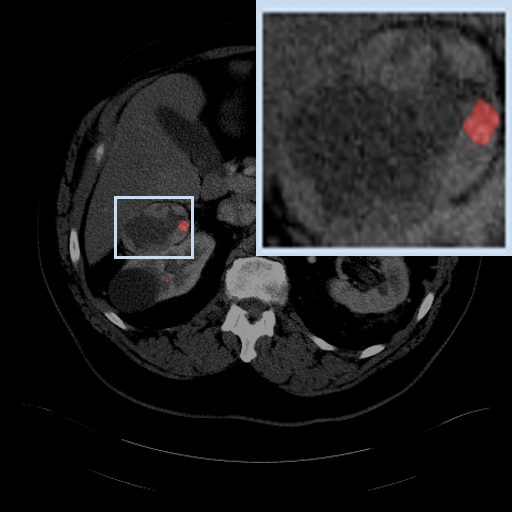}
                \end{minipage}
            }
            \subfigure{
                \begin{minipage}[t]{0.12\linewidth}
                    \centering
                    \includegraphics[width=1\linewidth]{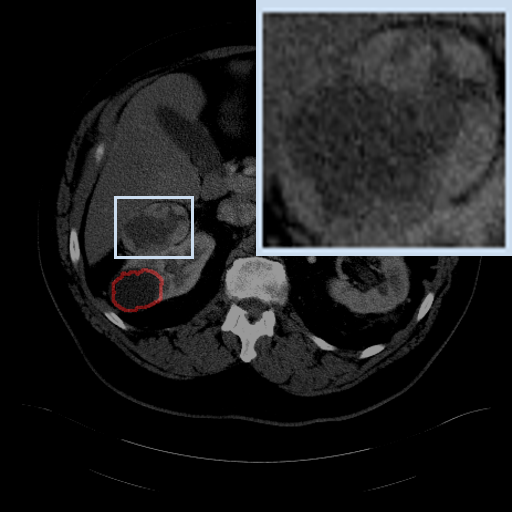}
                \end{minipage}
            }
        \end{minipage}
	
	\vspace{1mm}
 
        \begin{minipage}{\textwidth}
            \centering
            \rotatebox{90}{\scriptsize{~Pancreas-Case 200}}
            \subfigure{
                \begin{minipage}[t]{0.12\linewidth}
                    \centering
                    \includegraphics[width=1\linewidth]{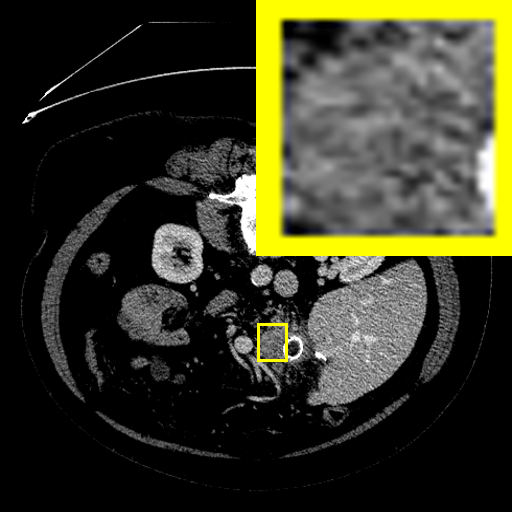}
                \end{minipage}
            }
            \subfigure{
                \begin{minipage}[t]{0.12\linewidth}
                    \centering
                    \includegraphics[width=1\linewidth]{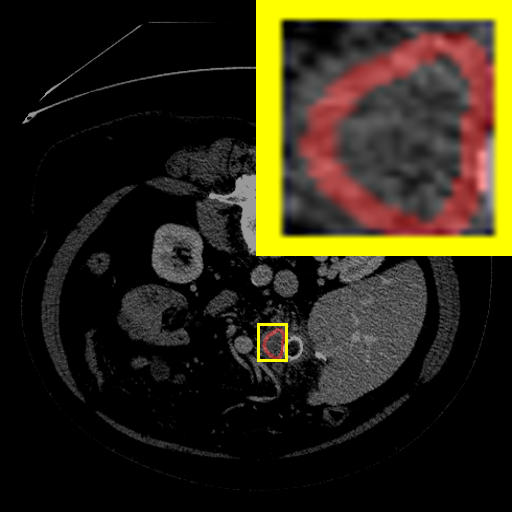}
                \end{minipage}
            }
            \subfigure{
                \begin{minipage}[t]{0.12\linewidth}
                    \centering
                    \includegraphics[width=1\linewidth]{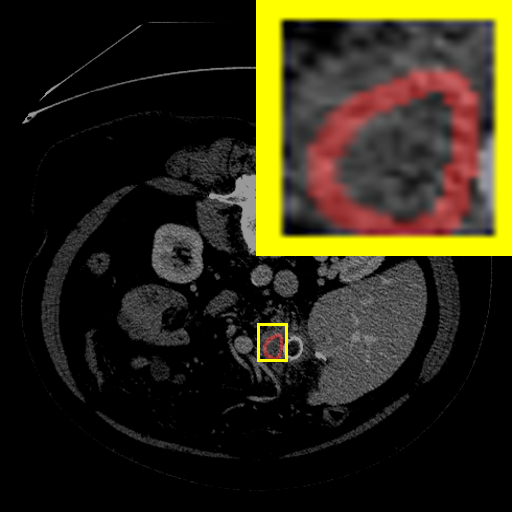}
                \end{minipage}
            }
            \subfigure{
                \begin{minipage}[t]{0.12\linewidth}
                    \centering
                    \includegraphics[width=1\linewidth]{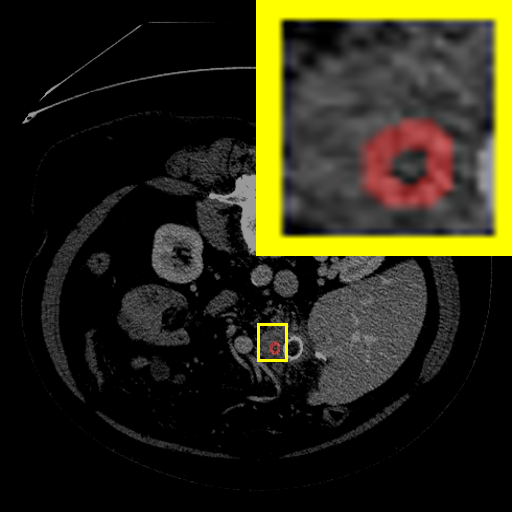}
                \end{minipage}
            }
            \subfigure{
                \begin{minipage}[t]{0.12\linewidth}
                    \centering
                    \includegraphics[width=1\linewidth]{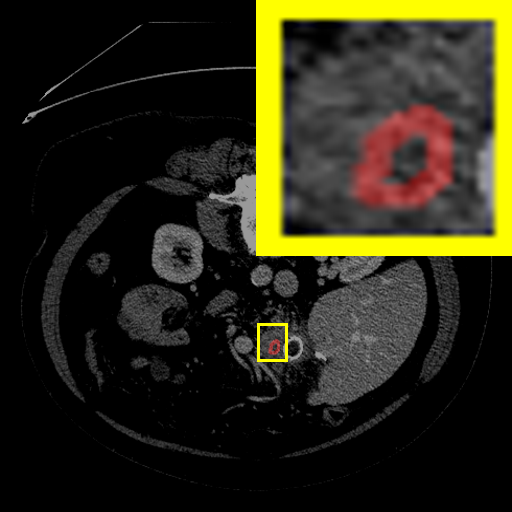}
                \end{minipage}
            }
            \subfigure{
                \begin{minipage}[t]{0.12\linewidth}
                    \centering
                    \includegraphics[width=1\linewidth]{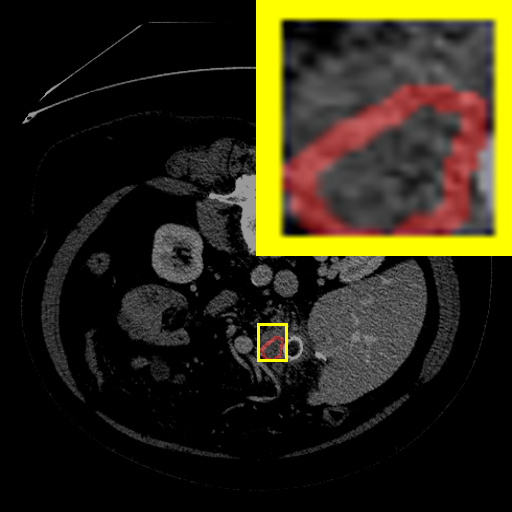}
                \end{minipage}
            }
            \subfigure{
                \begin{minipage}[t]{0.12\linewidth}
                    \centering
                    \includegraphics[width=1\linewidth]{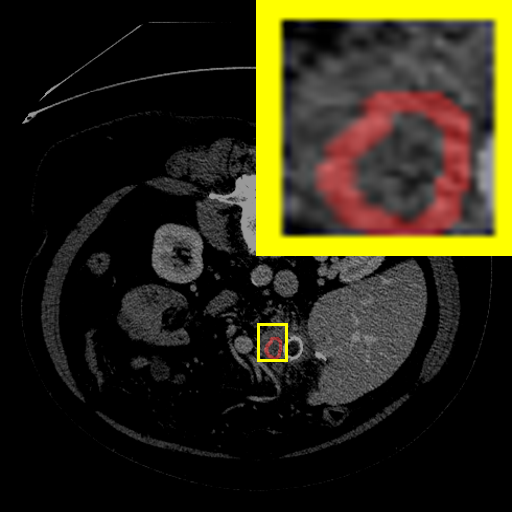}
                \end{minipage}
            }
        \end{minipage}
	
	\vspace{1mm}

         \begin{minipage}{\textwidth}
            \centering
            \rotatebox{90}{\scriptsize{~LiTS17-Case 29}}
            \subfigure{
                \begin{minipage}[t]{0.12\linewidth}
                    \centering
                    \includegraphics[width=1\linewidth]{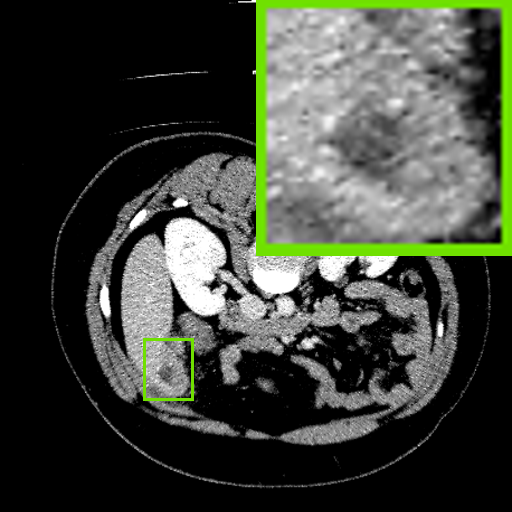}
                \end{minipage}
            }
            \subfigure{
                \begin{minipage}[t]{0.12\linewidth}
                    \centering
                    \includegraphics[width=1\linewidth]{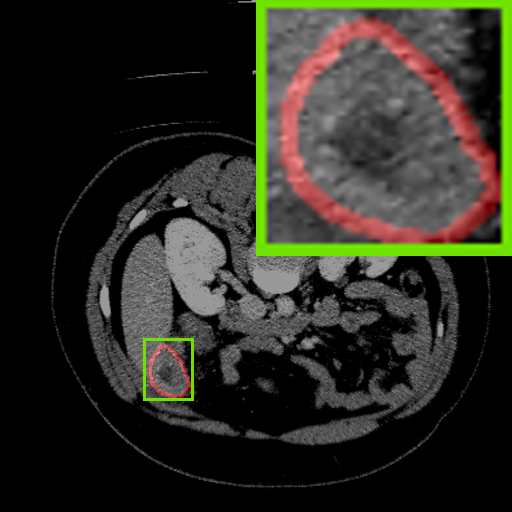}
                \end{minipage}
            }
            \subfigure{
                \begin{minipage}[t]{0.12\linewidth}
                    \centering
                    \includegraphics[width=1\linewidth]{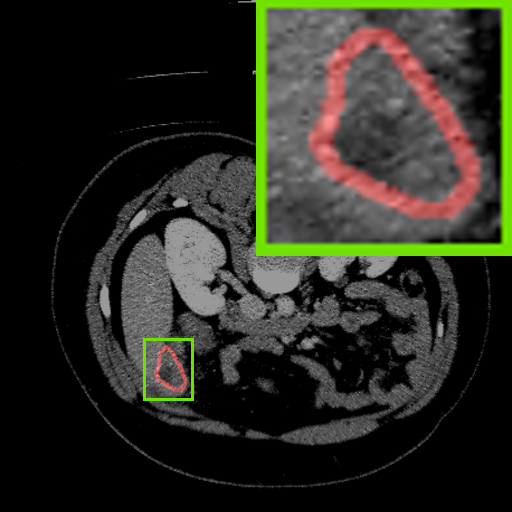}
                \end{minipage}
            }
            \subfigure{
                \begin{minipage}[t]{0.12\linewidth}
                    \centering
                    \includegraphics[width=1\linewidth]{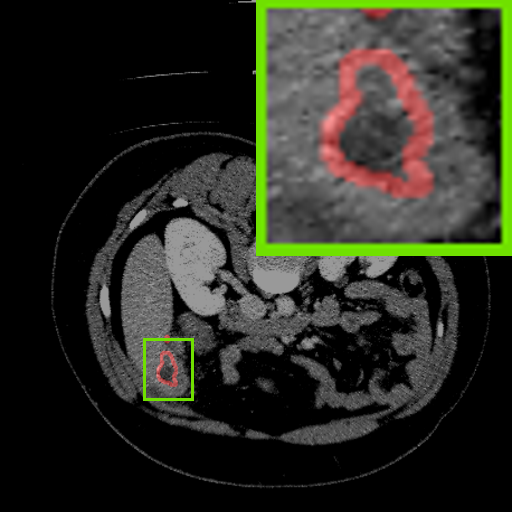}
                \end{minipage}
            }
            \subfigure{
                \begin{minipage}[t]{0.12\linewidth}
                    \centering
                    \includegraphics[width=1\linewidth]{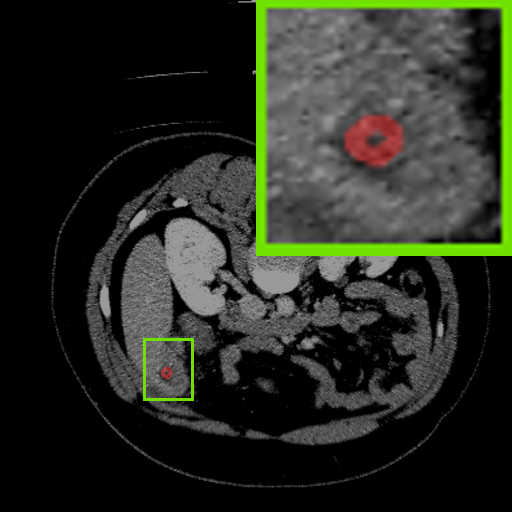}
                \end{minipage}
            }
            \subfigure{
                \begin{minipage}[t]{0.12\linewidth}
                    \centering
                    \includegraphics[width=1\linewidth]{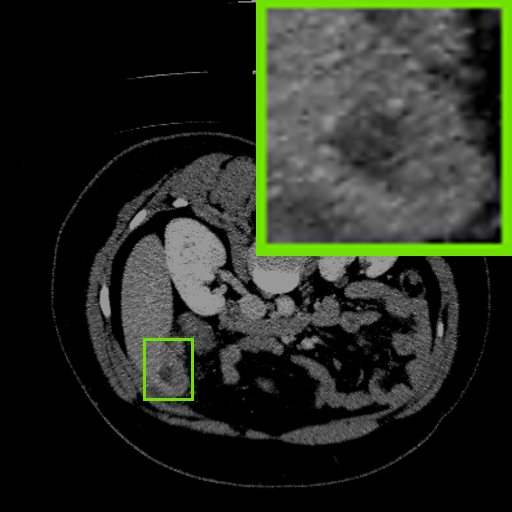}
                \end{minipage}
            }
            \subfigure{
                \begin{minipage}[t]{0.12\linewidth}
                    \centering
                    \includegraphics[width=1\linewidth]{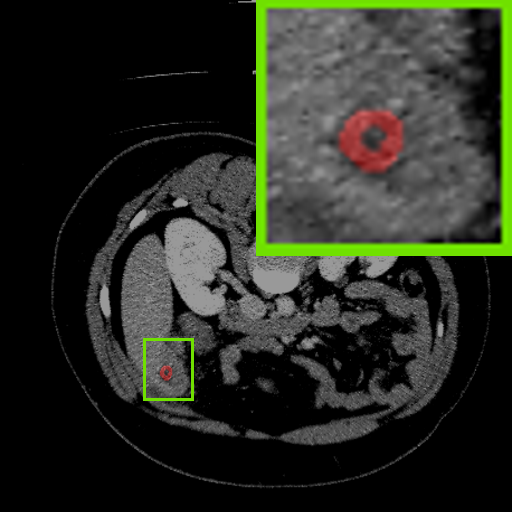}
                \end{minipage}
            }
        \end{minipage}
	
	\vspace{1mm}

        \setcounter{subfigure}{0}
        
        \begin{minipage}{\textwidth}
            \centering
            \rotatebox{90}{\scriptsize{~Colon-Case 9}}
            \subfigure[Image]{
                \begin{minipage}[t]{0.12\linewidth}
                    \centering
                    \includegraphics[width=1\linewidth]{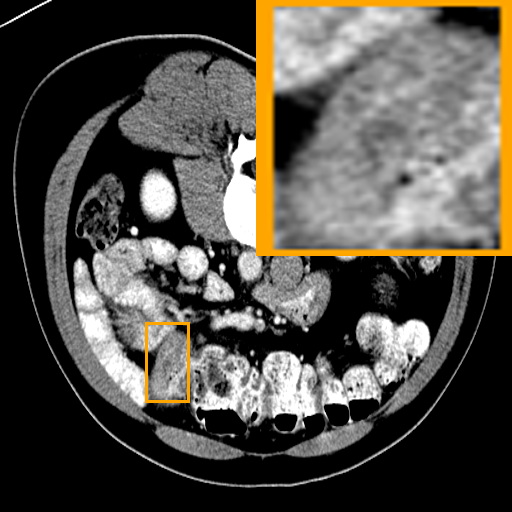}
                \end{minipage}
            }
            \subfigure[Ground Truth]{
                \begin{minipage}[t]{0.12\linewidth}
                    \centering
                    \includegraphics[width=1\linewidth]{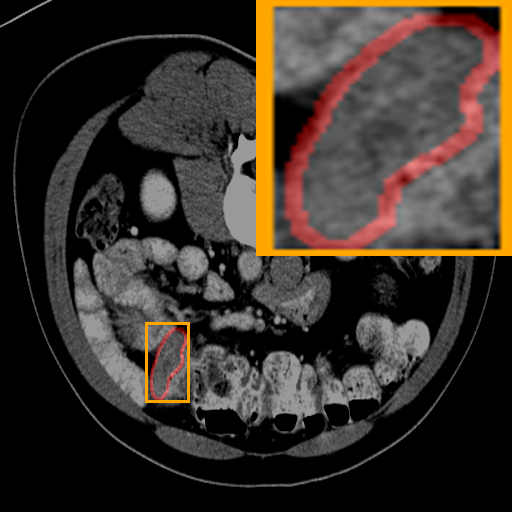}
                \end{minipage}
            }
            \subfigure[DEAP-3DSAM]{
                \begin{minipage}[t]{0.12\linewidth}
                    \centering
                    \includegraphics[width=1\linewidth]{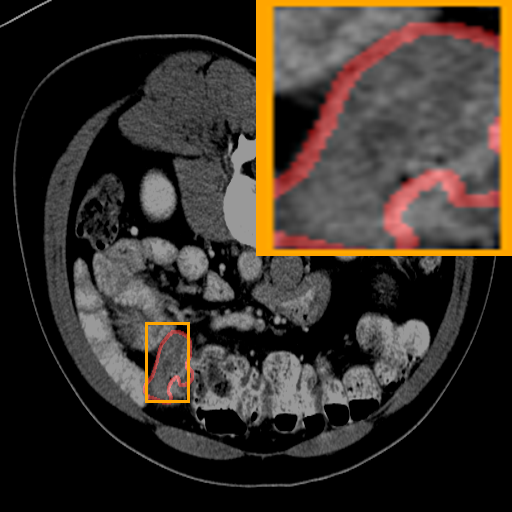}
                \end{minipage}
            }
            \subfigure[3DSAM-Adapter]{
                \begin{minipage}[t]{0.12\linewidth}
                    \centering
                    \includegraphics[width=1\linewidth]{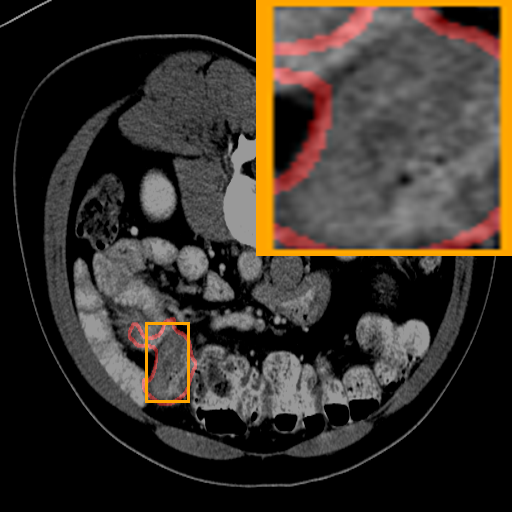}
                \end{minipage}
            }
            \subfigure[UNETR++]{
                \begin{minipage}[t]{0.12\linewidth}
                    \centering
                    \includegraphics[width=1\linewidth]{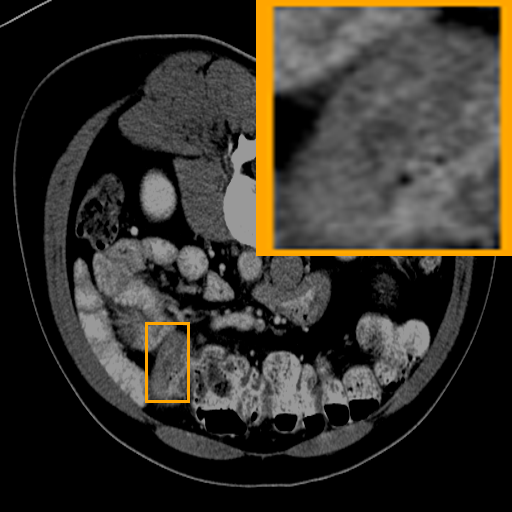}
                \end{minipage}
            }
            \subfigure[Swin-UNETR]{
                \begin{minipage}[t]{0.12\linewidth}
                    \centering
                    \includegraphics[width=1\linewidth]{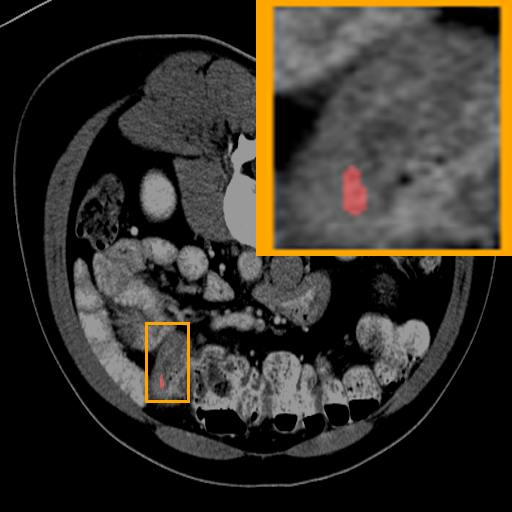}
                \end{minipage}
            }
            \subfigure[TransBTS]{
                \begin{minipage}[t]{0.12\linewidth}
                    \centering
                    \includegraphics[width=1\linewidth]{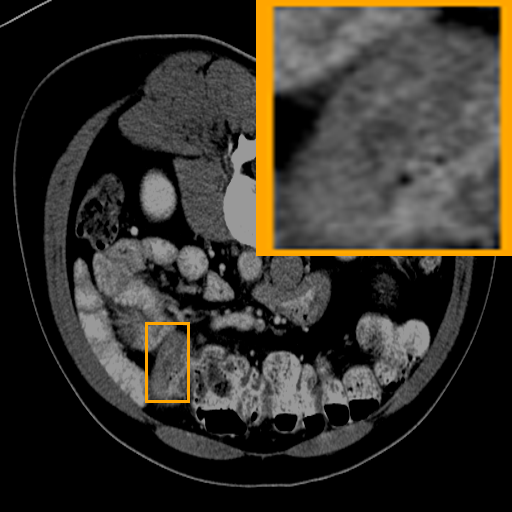}
                \end{minipage}
            }
        \end{minipage}
	\caption{Qualitative comparison visualization of DEAP-3DSAM and baselines on four datasets.}
	\label{overall visual}
\end{figure*}

\subsection{Dual Attention Prompter}
We introduce a unique dual attention mechanism serving as an automatic prompt encoder called the Dual Attention Prompter. As illustrated in Fig. \ref{framework} (c), the Dual Attention Prompter comprises Spatial Attention and Channel Attention. Spatial Attention is pixel-level spatial self-attention. To address the significant computational challenges, we utilize self-attention with linear complexity as proposed in Linformer \cite{linformer}. Specifically, we reduce the number of tokens belonging to $K_{i,SA}$ and $V_{i,SA}$ to a constant level through a linear layer. Overall, the calculation of Spatial Attention is as follows,
\begin{equation}
Q_{i,SA} = Norm(Z_iW_{i,SA}^q),  \\
\end{equation}
\begin{equation}
\hat{K}_{i,SA} = Linear(Z_iW_{i,SA}^k), \\
\end{equation}
\begin{equation}
\hat{V}_{i,SA} = Linear(Z_iW_{i,SA}^v), \\
\end{equation}
\begin{equation}
Z_{i,SA} = \hat{V}_{i,SA}Softmax(\hat{K}_{i,SA}^TQ_{i,SA}), \\
\end{equation}
where $W_{i,SA}^q, W_{i,SA}^k, W_{i,SA}^v \in \mathbb{R}^{C \times C}$ denote the weights for Spatial Attention; $Linear(\cdot)$ denotes the linear layer used to reduce the number of tokens to a constant $n$; $Norm(\cdot)$ denotes normalization layer; $Softmax(\cdot)$ denotes softmax function.

Channel Attention is designed to distill key information on channel aspects and enable the model focus on the most salient elements of channel features. Specifically, Channel Attention is computed as follows,
\begin{equation}
Q_{i,CA} = Norm(Z_iW_{i,CA}^q), \\
\end{equation}
\begin{equation}
K_{i,CA} = Norm(Z_iW_{i,CA}^k), \\
\end{equation}
\begin{equation}
V_{i,CA} = Z_iW_{i,CA}^v, \\
\end{equation}
\begin{equation}
Z_{i,CA} = V_{i,CA}Softmax(Q_{i,CA}^T K_{i,CA}), \\
\end{equation}
where $W_{i,CA}^q, W_{i,CA}^k, W_{i,CA}^v \in \mathbb{R}^{C \times C}$ denote the weights for Channel Attention. Inspired by Efficient Paired-Attention in UNETR++ \cite{unetr++}, we share $W_{i,SA}^q$ and $W_{i,SA}^k$ of Spatial Attention with Channel Attention, \ie $W_{i,SA}^q=W_{i,CA}^q, W_{i,SA}^k = W_{i,CA}^k$. It allows for better integrating the two attention modules and reduces computational complexity.

Finally, we concatenate the features from the two attentions and apply residual concatenation before output, as follows,
\beq
    \Tilde{Z}_i = Z_i + Concat(Z_{i,SA}W_{down, SA}, Z_{i,CA}W_{down,CA}),
\eeq
where $\Tilde{Z}_i$ denotes the feature map incorporating the automatic prompt information; $W_{down, SA}, W_{down, CA} \in \mathbb{R}^{C \times \frac{C}{2}}$ denote the linear mapping parameters. Note that we incorporate automatic prompt information only for $Z_{12}$. This results in the best performance, as shown in Table~\ref{prompt layer}.

\section{Experiment}

\subsection{Experiment Setup}
\subsubsection{Datasets}
To demonstrate the efficacy of our proposed DEAP-3DSAM for complex 3D medical segmentation tasks, we focus our experiments on four publicly available abdominal tumor datasets. Abdominal tumor segmentation is among the most challenging tasks in 3D medical segmentation. The details of these four datasets are provided below.
\begin{itemize}[leftmargin=*]
    \item \textbf{KiTS21}: Kidney Tumor Segmentation Challenge 2021, contains 300 abdominal CT scans.
    \item \textbf{Pancreas}: Pancreas Tumor Segmentation task from the 2018 MICCAI Medical Segmentation Decathlon Challenge, contains 281 abdominal CT scans.
    \item \textbf{LiTS17}: Liver Tumor Segmentation Challenge 2017, contains 118 abdominal CT scans.
    \item \textbf{Colon}: Colon Cancer Primaries Segmentation task from the 2018 MICCAI Medical Segmentation Decathlon Challenge, consists of 126 abdominal CT scans.
\end{itemize}

We adhere to the same settings as 3DSAM-Adapter \cite{3dsamadapter} reported in the paper, \ie only the tumor labels from the above datasets for our experiments. Each dataset is randomly divided into 70\%, 10\%, and 20\% for training, validation, and testing.

\subsubsection{Metrics}
We utilize the widely employed medical image segmentation metrics, the DICE score (DICE) and Normalized Surface Dice (NSD) \cite{nnsam, 3dsamadapter}, for performance evaluation. The DICE focuses on the overlap between the predicted regions and the ground truth regions, whereas the NSD emphasizes the alignment of the boundaries of these two regions.

\subsubsection{Implementation Details}
We implement our method on PyTorch. Moreover, we employ weighted dice loss and cross-entropy loss for supervised training, setting the weights to 0.5 and 0.5. We train our model with AdamW optimizer for a total of 200 epochs. We set the initial learning rate to $2e-4$ and the momentum to 0.9. We use SAM-B \cite{sam} as the pre-training weights for all SAM-based models. The other preprocessing and data augmentation are the same as in 3DSAM-Adapter \cite{3dsamadapter}. We adopt the average value of 5-fold cross-validation for all models and baselines as the authentic performance.

\begin{figure}[t]
	\centering
	
	\subfigure{
        \rotatebox{90}{\scriptsize{~KiTS21-Case 58}}
		\begin{minipage}[t]{0.2\linewidth}
			\centering
			\includegraphics[width=1\linewidth]{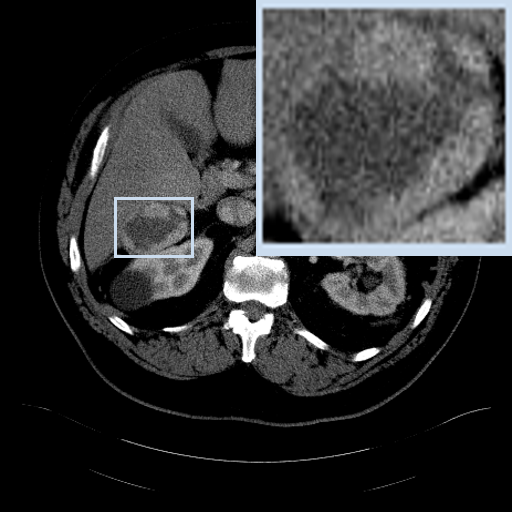}
		\end{minipage}
	}
	\subfigure{
		\begin{minipage}[t]{0.2\linewidth}
			\centering
			\includegraphics[width=1\linewidth]{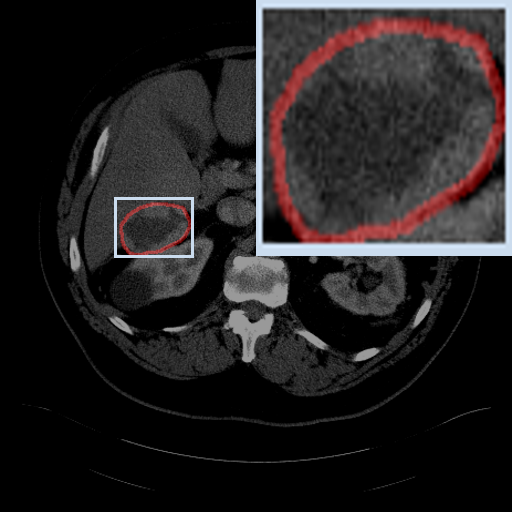}
		\end{minipage}
	}
	\subfigure{
		\begin{minipage}[t]{0.2\linewidth}
			\centering
			\includegraphics[width=1\linewidth]{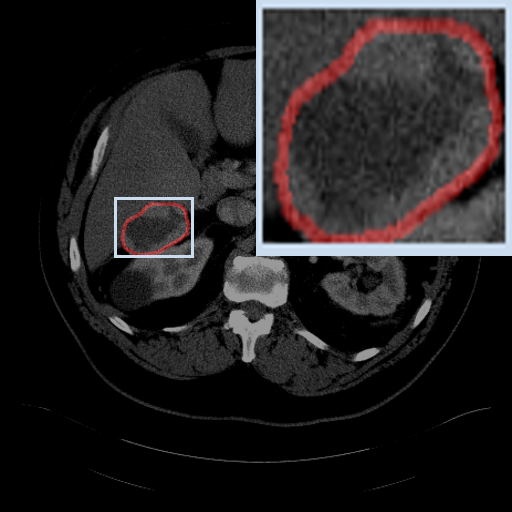}
		\end{minipage}
	}
	\subfigure{
		\begin{minipage}[t]{0.2\linewidth}
			\centering
			\includegraphics[width=1\linewidth]{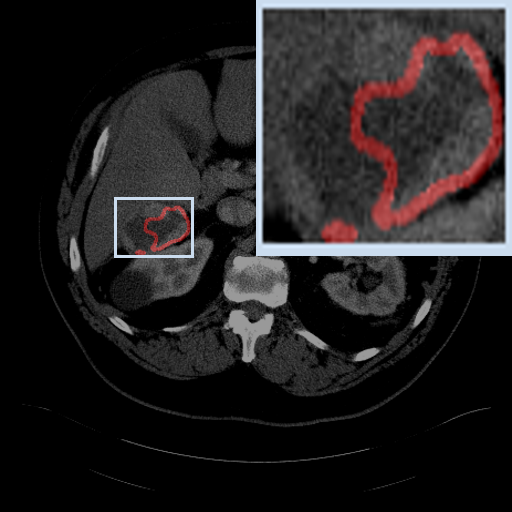}
		\end{minipage}
	}

	\vspace{-3mm}
	
	\subfigure{
        \rotatebox{90}{\scriptsize{Pancreas-Case 345}}
		\begin{minipage}[t]{0.2\linewidth}
			\centering
			\includegraphics[width=1\linewidth]{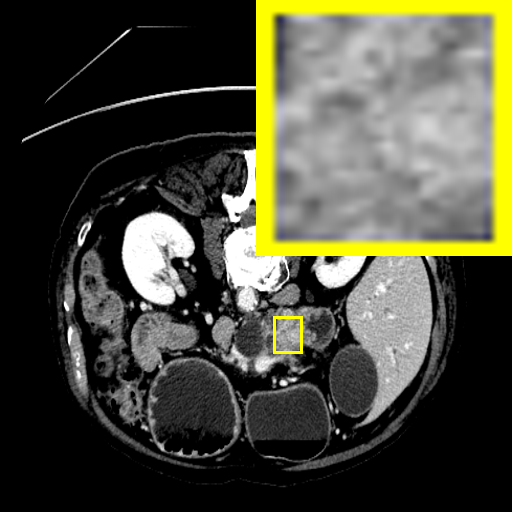}
		\end{minipage}
	}
	\subfigure{
		\begin{minipage}[t]{0.2\linewidth}
			\centering
			\includegraphics[width=1\linewidth]{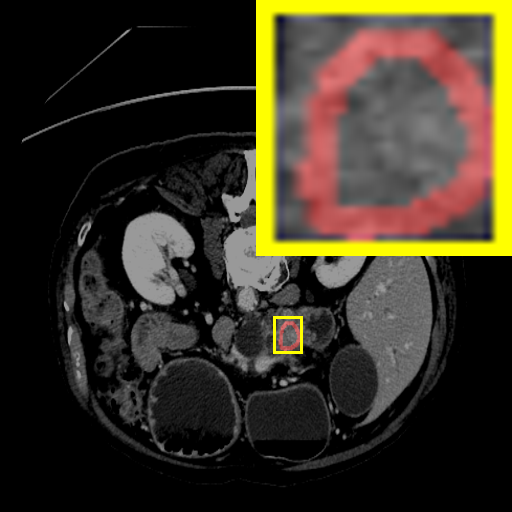}
		\end{minipage}
	}
	\subfigure{
		\begin{minipage}[t]{0.2\linewidth}
			\centering
			\includegraphics[width=1\linewidth]{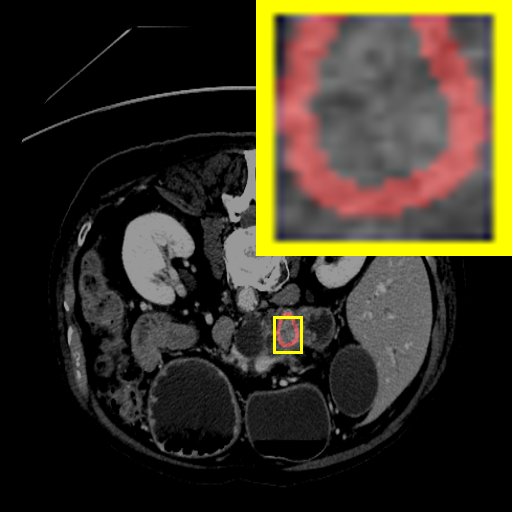}
		\end{minipage}
	}
	\subfigure{
		\begin{minipage}[t]{0.2\linewidth}
			\centering
			\includegraphics[width=1\linewidth]{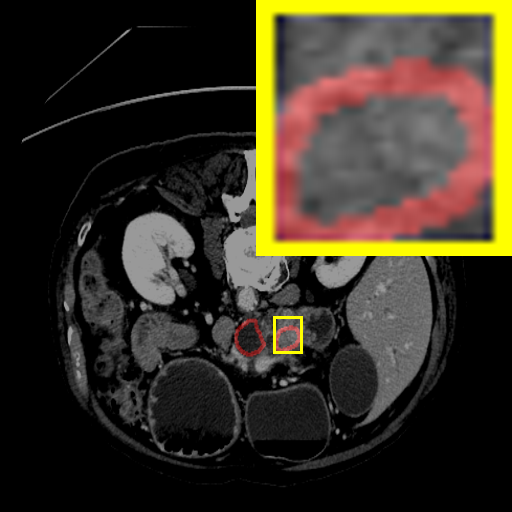}
		\end{minipage}
	}
	
	\vspace{-3mm}
    
	\subfigure{
        \rotatebox{90}{\scriptsize{~LiTS17-Case 76}}
		\begin{minipage}[t]{0.2\linewidth}
			\centering
			\includegraphics[width=1\linewidth]{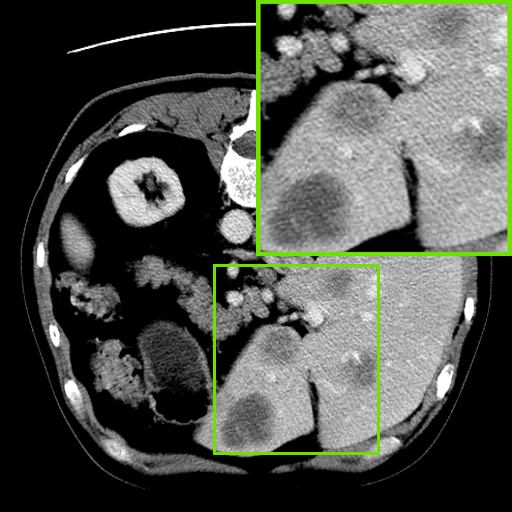}
		\end{minipage}
	}
	\subfigure{
		\begin{minipage}[t]{0.2\linewidth}
			\centering
			\includegraphics[width=1\linewidth]{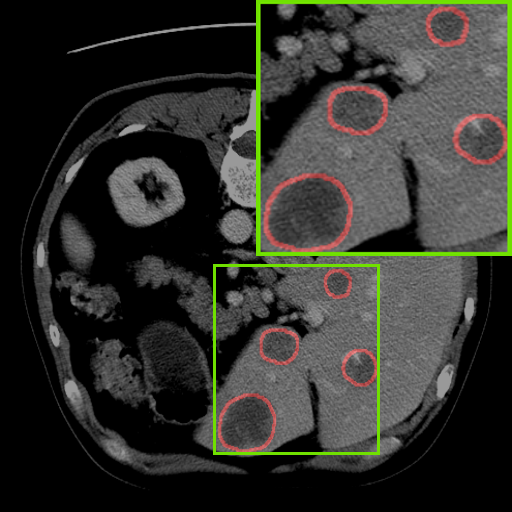}
		\end{minipage}
	}
	\subfigure{
		\begin{minipage}[t]{0.2\linewidth}
			\centering
			\includegraphics[width=1\linewidth]{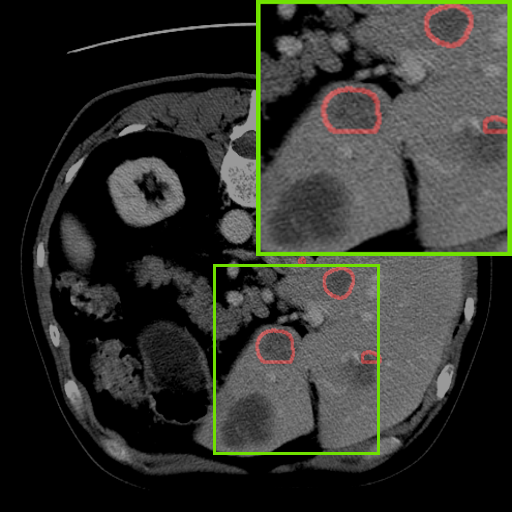}
		\end{minipage}
	}
	\subfigure{
		\begin{minipage}[t]{0.2\linewidth}
			\centering
			\includegraphics[width=1\linewidth]{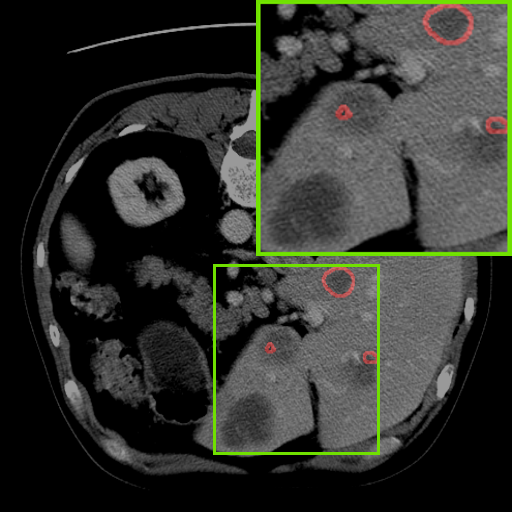}
		\end{minipage}
	}
	
	\vspace{-3mm}

	\setcounter{subfigure}{0}
        \subfigure[Image]{
		\rotatebox{90}{\scriptsize{~Colon-Case 141}}
		\begin{minipage}[t]{0.2\linewidth}
			\centering
			\includegraphics[width=1\linewidth]{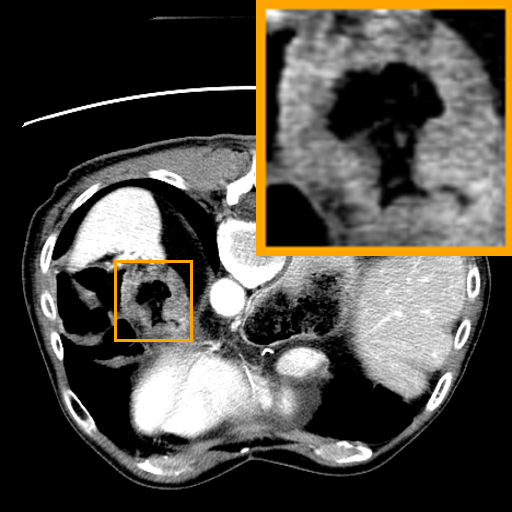}
		\end{minipage}
	}
	\subfigure[Ground Truth]{
		\begin{minipage}[t]{0.2\linewidth}
			\centering
			\includegraphics[width=1\linewidth]{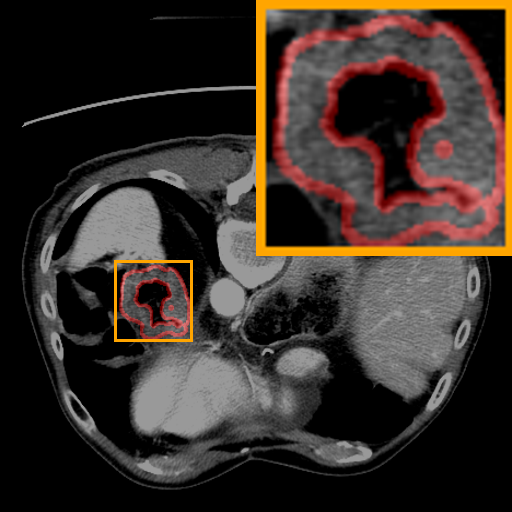}
		\end{minipage}
	}
	\subfigure[Ours]{
		\begin{minipage}[t]{0.2\linewidth}
			\centering
			\includegraphics[width=1\linewidth]{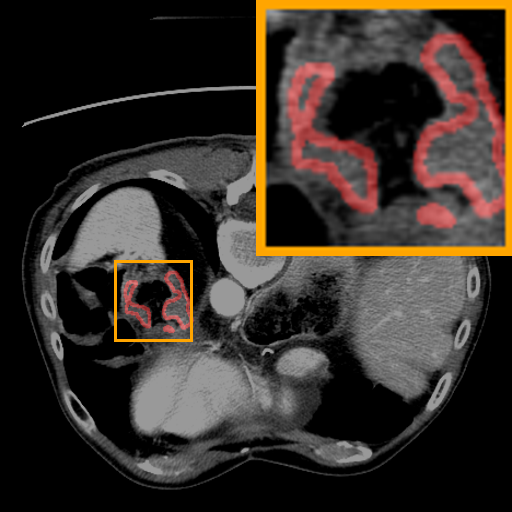}
		\end{minipage}
	}
	\subfigure[w/o FED]{
		\begin{minipage}[t]{0.2\linewidth}
			\centering
			\includegraphics[width=1\linewidth]{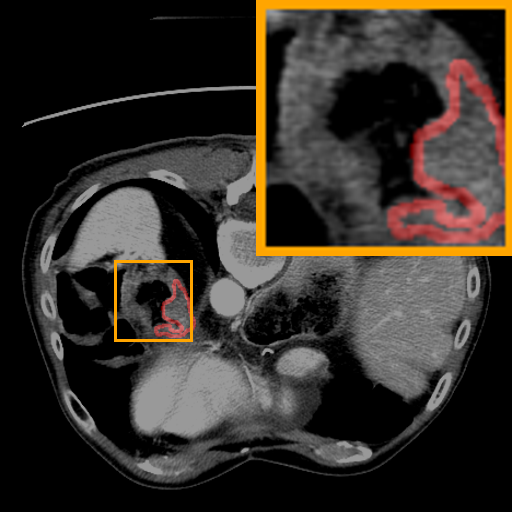}
		\end{minipage}
	}
	\caption{Qualitative analysis visualization of Feature Enhanced Decoder (FED) on four datasets.}
	\label{decoder ablation imgs}
\end{figure}

\begin{figure}[t]
  \centering
  \includegraphics[width=0.5\textwidth]{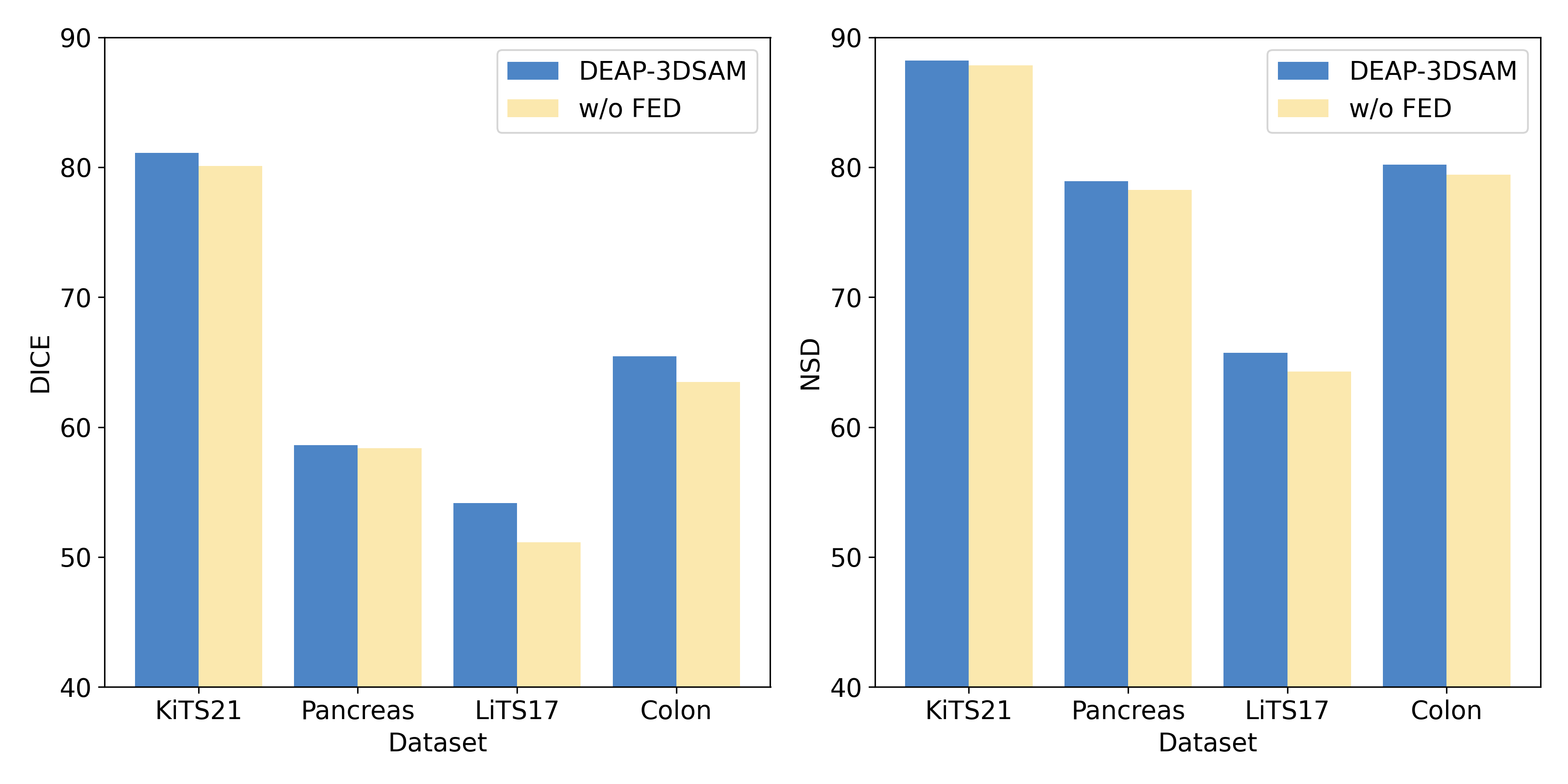}
  \caption{Quantitative analysis visualization of Feature Enhanced Decoder (FED) on four datasets.}
  \label{decoder ablation}
\end{figure}

\subsection{Comparison of Overall Performance}

We conduct a comprehensive comparison of the latest 3D medical image segmentation models, including state-of-the-art Transformer-based methods, including Swin-UNETR \cite{swinunetr}, UNETR++ \cite{unetr++}, TransBTS \cite{transbts}, nnFormer \cite{nnformer}, and CNN-based approaches like nnU-Net \cite{nnunet} and 3D UX-Net \cite{3duxnet}. Additionally, we evaluate the original SAM-B \cite{sam} model. Specifically, we slice the 3D medical images by depth and given 10 point prompts per slice, denoted as SAM-B (10 pts/slice). Furthermore, we compare the SAM-based model, 3DSAM-Adapter \cite{3dsamadapter}, which also employs pseudo 3D patching. We assess performance across three scenarios: 1 point, 3 points, and 10 points, denoted as 3DSAM-Adapter (1 pt/volume), 3DSAM-Adapter (3 pts/volume), and 3DSAM-Adapter (10 pts/volume). Note that we reset the patch size to (128, 128, 128) for all datasets when training 3DSAM-Adapter \cite{3dsamadapter}.

\subsubsection{Quantitative Performance Comparison} According to the results presented in Table~\ref{overall performance}, we have made several observations. Firstly, with Feature Enhanced Decoder and Dual Attention Prompter, DEAP-3DSAM\ddag ~and DEAP-3DSAM\dag ~achieve state-of-the-art and sub-optimal performance on the KiTS21 and Colon datasets. This demonstrates the effectiveness and advancement of DEAP-3DSAM in 3D medical image segmentation. Furthermore, DEAP-3DSAM outperforms the 3DSAM-Adapter (10 pts/volume) \cite{3duxnet} on the LiTS17 dataset and the 3DSAM-Adapter (1 pt/volume) on the Pancreas dataset. This indicates that DEAP-3DSAM offers performance comparable to that of interactive segmentation models. Secondly, the best SAM-based models surpass traditional CNN- or Transformer-based models across three datasets: KiTS21, Pancreas, and Colon. Notably, on the Colon dataset, the optimal SAM-based model achieves a 49.02\% increase in the DICE score and a 52.73\% increase in the NSD score. This confirms the effectiveness of SAM-based methods in 3D medical image segmentation and highlights their potential. Lastly, despite the adequate number of prompts, the performance of SAM-B \cite{sam} with 10 prompts per slice is unsatisfactory. This suggests that SAM possesses considerable potential, but its adaptation to 3D medical image segmentation requires a suitably designed structure. Additionally, the parameter count of DEAP-3DSAM is comparable to that of conventional methods, underscoring its efficiency.

It is noteworthy that the performance of DEAP-3DSAM and 3DSAM-Adapter \cite{3dsamadapter} on LiTS17 is less robust than that of nnU-Net \cite{nnunet}, and this warrants further exploration in the future. As is well known, liver tumors often consist of multiple smaller tumors that are scattered throughout the liver. One plausible explanation is that SAM-based methods may be less effective at addressing targets with a dispersed distribution.

\subsubsection{Qualitative Performance Comparison} We also performed qualitative analysis on four datasets. As illustrated in Fig.~\ref{overall visual}, DEAP-3DSAM accurately identifies the target regions and closely matches their size. In contrast, 3DSAM-Adapter \cite{3dsamadapter} exhibits limitations in matching the size and shape of the target regions. This proves that DEAP-3DSAM captures more complex image features, owing to its Dual Attention Prompter and Feature Enhanced Decoder. Furthermore, while these SAM-based methods are nearly capable of localizing the target regions, many traditional methods, \ie UNETR++ \cite{unetr++}, Swin-UNETR \cite{swinunetr}, and TransBTS \cite{transbts}, struggle to achieve this. This highlights the potential of SAM-based methods for addressing complex 3D segmentation tasks.

\begin{table}[t]
    \centering
    \caption{Effects of different prompt layers on Colon. }
    \begin{tabular}{l|l|l|l|l|l|l}
    \hline
        \textbf{Prompt Layer} & \textbf{\text{3-th}} & \textbf{\text{6-th}} & \textbf{\text{9-th}} & \textbf{\text{12-th}} & \textbf{DICE}  $\uparrow$ & \textbf{NSD} $\uparrow$ \\ \hline
        For 3-th Layer & \Checkmark & ~ & ~ & ~ & 65.16 & 79.95 \\ 
        For 6-th Layer & ~ & \Checkmark & ~ & ~ & 62.74 & 77.24 \\ 
        For 9-th Layer & ~ & ~ & \Checkmark & ~ & 62.94 & 77.77 \\ 
        For 12-th Layer & ~ & ~ & ~ & \Checkmark & \textbf{65.43} & \textbf{80.21} \\ \hline
    \end{tabular}
    \label{prompt layer}
\end{table}

\begin{table}[t]
    \centering
    \caption{Effects of different prompt methods on LiTS17 and Colon.}
    \begin{tabular}{l|cc|cc}
    \hline
        \multirow{2}*{\textbf{Models}} & \multicolumn{2}{c|}{\textbf{LiTS17}} & \multicolumn{2}{c}{\textbf{Colon}}   \\  \cline{2-5}
        ~ & \textbf{DICE} $\uparrow$ & \textbf{NSD} $\uparrow$ & \textbf{DICE} $\uparrow$ & \textbf{NSD} $\uparrow$ \\ \hline
        DEAP-3DSAM (Base) & \textbf{54.16} & 65.71 & \textbf{65.43} & \textbf{80.21} \\ \hline
        - Channel Attn & 51.97 & 64.08 & 64.49 & 79.47 \\ \hline
        - Spatial Attn (No Prompt) & 51.69 & 63.65 & 63.04 & 78.69 \\  \hline
        + Points (1 pt/volume) & 51.95 & 64.12 & 63.98 & 78.31 \\ \hline
        + Points (10 pts/volume) & 53.64 & \textbf{66.31} & 65.19 & 79.51 \\ \hline
    \end{tabular}
    \label{prompt ablation}
\end{table}

\begin{table}[t]
    \centering
    \caption{Computation cost comparison of different prompt methods. The shape of the input feature map is $32 \times 32 \times 32 \times 256$.}
    \begin{tabular}{l|l|l}
    \hline
        \textbf{Prompt Method} & \textbf{FLOPs} & \textbf{Params} \\ \hline
        Point Prompt & 4.41G & 2.90M \\ 
        Spatial Attention Prompter & 9.66G & 4.46M \\ 
        Dual Attention Prompter & 11.81G & 4.52M \\ 
        Full Dual Attention Prompter & 16.17G & 4.65M \\ \hline
    \end{tabular}
    \label{complexity}
\end{table}

\subsection{Effects of Feature Enhanced Decoder}

In order to demonstrate the effectiveness of our proposed Feature Enhanced Decoder, we conducted comprehensive experiments on four datasets, evaluating both quantitative and qualitative aspects.

\subsubsection{Quantitative analysis of Feature Enhanced Decoder} The results are presented in Fig. \ref{decoder ablation}. Note that ``w/o FED'' is a variant of DEAP-3DSAM that utilizes a basic Multi-Layer Aggregation (MLA) decoder. As shown, the performance of DEAP-3DSAM declines when the original image features are removed across four abdominal tumor datasets. Notably, in the LiTS17 dataset, the DICE and NSD metrics decreased by 5.55\% and 2.17\%, respectively. These findings highlight the substantial influence of the Feature Enhanced Decoder on enhancing segmentation performance.

\subsubsection{Qualitative analysis of Feature Enhanced Decoder} We also perform qualitative analyses on several cases, as illustrated in Fig. \ref{decoder ablation imgs}. The visualizations indicate that removing the original image features leads to a significant inaccuracy in the size of the segmented regions across the four datasets. This outcome underscores the importance of the original image features in accurately capturing the size of the target region. This aligns with our intuition that incorporating the original image information enhances the spatial features, resulting in more precise segmentation in both size and shape.

\subsection{Effects of Dual Attention Prompter}

In order to thoroughly investigate the impact of the Dual Attention Prompter, we present three aspects of the analysis: the effects of different prompt layers, the performance comparison of various prompting methods, and the comparison of computational costs.

\subsubsection{Effects of different prompt layers} To optimize the application of the Dual Attention Prompter, we investigate the effects of positioning it at different layers on the Colon dataset. According to the results presented in Table~\ref{prompt layer}, the automatic prompt information at the 3-th and 12-th layers is more effective in improving the segmentation performance. The performance improvement observed in the intermediate layers, such as the 6-th and 9-th layers, is minimal. This discrepancy may arise from the ability of shallower layers to retain spatial details, which allows Spatial Attention to capture broader contextual cues. In contrast, the deeper layers, rich in channel data, enable Channel Attention to explore more intricate semantic features.

\subsubsection{Effects of different prompt methods} Table \ref{prompt ablation} illustrates the performance of various prompting methods. In the experimental setup, we either remove (indicated by ``-'') or add (indicated by ``+'') a module from top to bottom to modify the prompting methods, using DEAP-3DSAM as the base model. For example, ``+ Points (1 pt/volume)'' signifies the complete removal of the Dual Attention Prompter while adding one point prompt per volume. Note that the point prompting technique, initially introduced in SEEM \cite{seem}, has become a widely adopted approach in 3D medical imaging for SAM \cite{3dsamadapter, promise}. The results indicate that removing Channel Attention and Spatial Attention leads to a gradual decline in performance, underscoring the effectiveness of the Dual Attention Prompter. Even with the addition of 10 points per volume, the performance remains inferior to that of the base DEAP-3DSAM model, demonstrating that our proposed Dual Attention Prompter possesses superior prompting capacity compared to existing manual prompting methods.

\subsubsection{Computation Cost Comparison} In real-world scenarios, the efficiency of segmentation methods is crucial. We quantify the temporal and spatial complexity of various prompting methods, evaluated through the metrics of Floating Point Operations (FLOPs) and Parameter count (Params), respectively. The evaluation results are presented in Table~\ref{complexity}, where ``Full Dual Attention Prompter'' means no weight sharing between Spatial Attention and Channel Attention. Our findings suggest that Spatial Attention does not generate a significant number of parameters or computations, thanks to Linear self-attention \cite{linformer}. Furthermore, the parameter sharing between Spatial Attention and Channel Attention leads to a 27\% reduction in FLOPs without any significant degradation in performance, as shown in Table~\ref{overall performance}. Overall, the number of parameters and computations for the Dual Attention Prompter is still acceptable compared to the point prompting.

\section{Conclusion}
In this work, we propose the DEAP-3DSAM architecture, an innovative \textbf{D}ecoder \textbf{E}nhanced and \textbf{A}uto \textbf{P}rompt \textbf{SAM} designed for \textbf{3D} medical image segmentation. Our DEAP-3DSAM first integrates original image features through a Feature Enhanced Decoder, thereby enriching the spatial feature representation. Secondly, our DEAP-3DSAM incorporates the Dual Attention Prompter, which leverages Spatial Attention and Channel Attention to autonomously identify relevant prompt information within feature maps. Our comprehensive experimental evaluation across four abdominal tumor segmentation datasets, \ie KiTS21, Pancreas, LiTS17, and Colon, indicates that our DEAP-3DSAM delivers performance on par with or superior to existing manual prompt techniques. Furthermore, we substantiate the efficacy of our Feature Enhanced Decoder and Dual Attention Prompter through rigorous ablation studies. Consequently, our DEAP-3DSAM can effectively perform 3D medical image segmentation without manual prompting, making it suitable for future automated medical-assisted diagnosis. It can be utilized not only for the automatic localization of lesion areas but also for extracting medical image features for disease prediction. Although our DEAP-3DSAM has not achieved the best performance on LiTS17, we are dedicated to improving SAM-based models for dispersed targets in future research.

\section*{Ackowledgment}
This work is supported by the Hunan Provincial Natural Science Foundation (Grant No. 2022JJ30668), the National High Level Hospital Clinical Research Funding (2023-PUMCH-E-006), and the National Key Research and Development Project of China (No. 2021ZD0110700).

\bibliographystyle{IEEEtran}
\bibliography{reference}  

\end{document}